\documentclass{article}

\usepackage[preprint]{neurips_2021}

\usepackage{microtype}
\usepackage{graphicx}
\usepackage{subfigure}
\usepackage{url}            
\usepackage{booktabs} 
\usepackage{xcolor}         
\usepackage[ruled,vlined]{algorithm2e}

\setcitestyle{square,numbers}

\usepackage{multicol}
\usepackage{amssymb}
\usepackage[utf8]{inputenc} 
\usepackage[T1]{fontenc}    
\usepackage{amsfonts}       
\usepackage{nicefrac}       
\usepackage{microtype}      
\usepackage[usecolors=true]{jlcode}    
\usepackage{float}
\usepackage{listings}
\newfloat{lstfloat}{htbp}{lop}
\floatname{lstfloat}{Listing}

\usepackage{hyperref}

\title{Designing Machine Learning Pipeline Toolkit for AutoML Surrogate Modeling Optimization}
\author{%
  Paulito P.~Palmes\\
  IBM Research \\
  Mulhuddart, Dublin 15, Ireland \\
  \texttt{paulpalmes@ie.ibm.com} \\
  \And
  Akihiro Kishimoto\\
  IBM Research \\
  Mulhuddart, Dublin 15, Ireland \\
  \texttt{akihirok@ie.ibm.com}
  \And
  Radu Marinescu\\
  IBM Research \\
  Mulhuddart, Dublin 15, Ireland \\
  \texttt{radu.marinescu@ie.ibm.com}\\
  \And 
  Parikshit Ram\\
  IBM Research\\
  Sandy Springs, GA, United States\\
  \texttt{parikshit.ram@ibm.com}\\
  \And
  Elizabeth Daly\\
  IBM Research \\
  Mulhuddart, Dublin 15, Ireland \\
  \texttt{elizabeth.daly@ie.ibm.com}
}

\begin{document}
\maketitle

\begin{abstract}
  The pipeline optimization problem in machine learning requires simultaneous
  optimization of pipeline structures and parameter adaptation of their
  elements. Having an elegant way to express these structures can help lessen
  the complexity in the management and analysis of their performances together
  with the different choices of optimization strategies. With these issues in
  mind, we created the AutoMLPipeline (AMLP) toolkit which facilitates the creation and
  evaluation of complex machine learning pipeline structures using simple
  expressions.  We use AMLP to find optimal pipeline signatures, 
  datamine them, and use these datamined features to speed-up learning
  and prediction. We formulated a two-stage pipeline optimization 
  with surrogate modeling in AMLP which outperforms other AutoML 
  approaches with a 4-hour time budget in less than 5 minutes of AMLP
  computation time. 
\end{abstract}

\section{Introduction}
\label{intro}

The typical machine learning (ML) workflow for classification and prediction
requires some or a combination of the following preprocessing steps:
\begin{itemize}
\item \emph{data cleaning}: Imputation, Interpolation, etc.
\item \emph{feature transformation}: Normalization, Scaling, One-hot encoding, etc.
\item \emph{feature selection}: Anova, Correlation, etc.
\item \emph{feature extraction}: PCA, ICA, FactorAnalysis, etc.
\item \emph{modeling}: RandomForest, XGBoost, SVM, AdaBoost, etc.
\end{itemize}

Each step has several choices of functions to use, together with their
corresponding parameters to initialize. Optimizing the performance of the
entire pipeline is a combinatorial search for the proper order and combination
of preprocessing steps, optimization of their corresponding parameters,
and a search for the optimal model and its hyper-parameters.

Because of close dependencies among these various steps to perform
optimization, the entire process is commonly called combined algorithm
selection and hyper-parameter
optimization or CASH~\cite{feurer2015askl,Thornton:KDD2013}.  Having an elegant way to
express pipeline structures can help lessen the complexity in the management
and analysis of the wide-array of choices of optimization routines. 
AMLP (\url{https://github.com/IBM/AutoMLPipeline.jl}) 
toolkit aims to address  this issue by supporting the following
features:
\begin{itemize}
\item Pipeline API that allows high-level description of modeling and 
      preprocessing workflow to support explainability and easy 
      dataminining of optimal pipeline signatures
\item Symbolic pipeline parsing to facilitate easy expression 
      of complex pipeline structures
\item Common API wrappers for ML libs including
      \emph{scikit-learn}~\cite{pedregosa2011},
      \emph{caret}~\cite{caret2008}, etc.
\item Easily extensible architecture by
      overloading two interfaces: \texttt{fit!} and \texttt{transform!}
\item High-level implementation of meta-ensembles
      for the ensemble composition of ensembles (recursively if needed) for robust prediction routines.
\item Categorical and numerical feature selectors
      for specialized preprocessing routines based on types.
\item High-level code of parallelized cross-validation (CV)
      routines by leveraging on the multi-threading and
      distributed computing features
      in \emph{Julia}~\cite{bezanson2017julia}.
\end{itemize}

AMLP's main feature is the use of relatively compact symbolic expression in
pipeline composition.  For instance, a pipeline expression to extract the
numerical features (\texttt{numf}) for 
\texttt{pca} decomposition, concatenated with one-hot
encoding (\texttt{ohe}) of categorical features (\texttt{catf})
of a given data for RandomForest (\texttt{rf}) modeling can be written
as:
\begin{lstlisting}[language=Julia,numbers=none]
pipe = @pipeline (numf |> pca) + (catf |> ohe) |> rf 
\end{lstlisting}

AMLP is written in \emph{Julia} language \cite{bezanson2017julia} to leverage on the
latter's support of modern features such as: multiple dispatch, just-in-time
(\emph{JIT}) compilation, multi-threading, parallel and distributed computing,
dynamic types, coroutines, interactive shell, high-performance, code specialization,
metaprogramming, and type inference.  Pure \emph{Julia} ML models in AMLP is
easy to maintain and extend by relying on just one programming language
because Julia's \emph{JIT} avoids the need to implement some
performance-critical tasks in C/C++.

\subsection{Related Work}
\label{relatedwork}

AMLP's main design was inspired by the
\emph{Orchestra}~\cite{orchestra2014} package.  Early ML packages for
\emph{Julia} are wrappers of existing ML toolkits from \emph{scikit-learn}  in
\emph{Python}~\cite{python95} and \emph{caret} in \emph{R}~\cite{r13}.  By
having common APIs across different ML implementations, \emph{Orchestra} showed
that one can mix and match ML preprocessing routines from \emph{caret},
\emph{scikit-learn}, and \emph{Julia} seamlessly in a pipeline by leveraging on
\emph{PyCall}~\cite{pycall} and \emph{RCall}~\cite{rcall} wrappers for
\emph{Python} and \emph{R}, respectively. This is a convenient package
because there are many cases where implementations of ML functions can be found
exclusively in either \emph{scikit-learn} or \emph{caret}.  Having both
libraries available for perusal is a great productivity boost for machine
learners in \emph{Julia}. Unfortunately, \emph{Orchestra} package has not been
maintained for more than 6 years, but its legacy lives in the AMLP's
design. Another popular toolkit in \emph{Julia} ML ecosystem is the
\emph{MLJ}~\cite{anthony_blaom_2019_3541506} package developed at the Alan Turing
Institute. It serves as a meta-package of ML libraries both native to \emph{Julia} as
well as \emph{scikit-learn}.

IBM~\footnote{IBM and the IBM logo are trademarks of International Business Machines Corporation, registered in many jurisdictions worldwide. Other product and service names might be trademarks of IBM or other companies. A current list of IBM trademarks is available on \url{http://ibm.com/trademark}.} \emph{Lale}~\cite{lale} is a toolkit in \emph{Python} with sophisticated
support of pipeline optimization useful for AutoML~\cite{Liu:AAAI2020}
algorithm development.  \emph{Lale} builds on \emph{scikit-learn} and
extends it by supporting features such as: automation, correctness checks, and
interoperability.  \emph{Lale} has consistent high-level interface for popular
AutoML algorithms such as:
\emph{Hyperopt}~\cite{bergstra2011nips,bergstra2012},
\emph{GridSearchCV}~\cite{pedregosa2011}, and
\emph{SMAC}~\cite{hutter2011smac}.  \emph{Lale} makes it possible to have an
easy comparison of the different AutoML algorithms for benchmarking, and research,
as well as for applications. While AMLP is at the early stage of development, its
pipeline architecture can be used as building blocks for the development of
future AutoML algorithms in \emph{Julia}'s ecosystem. For instance, we were 
able to create a \emph{Julia} wrapper of \emph{Lale} (\url{https://github.com/IBM/Lale.jl}) 
in less than a week of work by adapting and extending AMLP's type 
hierarchy and interfaces using multiple dispatch and operator overloading to 
emulate the pipeline grammar in \emph{Lale}.

\section{AMLP Architecture}
\label{amlparchitecture}

The code in Listing \ref{lst:abstypes} describes the abstract type hierarchy
used in the AMLP toolkit. At the top of the hierarchy is the \texttt{Machine}
abstraction with abstract functions \texttt{fit!} and \texttt{transform!} to be
implemented by its subtypes. Calling both functions in sequence are handled by
\texttt{fit\_transform!}.  A \texttt{Machine} has two subtypes:
\texttt{Computer} and \texttt{Workflow}.  A \texttt{Computer} can either be a
\texttt{Transformer} or a \texttt{Learner} while a \texttt{Workflow} can be
either a \texttt{Pipeline} or  \texttt{ComboPipeline}.

\begin{lstfloat}[htbp]
\begin{lstlisting}
abstract type Computer    <: Machine  
abstract type Workflow    <: Machine  
abstract type Learner     <: Computer  
abstract type Transformer <: Computer  

function fit!(mc::Machine, input::DataFrame, output::Vector)
   error(typeof(mc)," has no implementation.")
end

function transform!(mc::Machine, input::DataFrame)
   error(typeof(mc)," has no implementation.")
end

function fit_transform!(mc::Machine, input::DataFrame, output::Vector)
   fit!(mc,input,output)
   transform!(mc,input)
end


\end{lstlisting}
  \caption{Abstract Type Hierarchy}
  \label{lst:abstypes}
\end{lstfloat}

All \texttt{Machine} subtypes are expected to define their own \texttt{fit!}
and \texttt{transform!}  before they can be part of the elements in a
\texttt{Workflow}. Any \texttt{Workflow} instance uses these two functions
during training, feature transformation, prediction,
and cross-validation (CV). By
convention in \emph{Julia}~\cite{bezanson2017julia}, functions ending
with exclamation mark (\texttt{!}) mutate the value(s) of their arguments which
is done by \texttt{fit!} and \texttt{transform!}
to mutate the states and parameters of the
\texttt{Machine} instance in their arguments.

The AMLP workflow is based on the \emph{Orchestra}~\cite{orchestra2014}
package which drew its inspiration from the \emph{Unix} pipeline~\cite{unix84}.
The main elements of a pipeline are a series of \texttt{Computer} instances with each
instance performing a specific task. A typical pipeline for classification or
prediction  contains a series of transformers terminated by a learner.  During
\texttt{fit\_transform!}, these series of transformers act as filters
converting the input data into the same mathematical or statistical transform
before feeding them to the learner for training and prediction. In a pipeline
expression where the last element is not a learner, \texttt{fit\_transform!}
acts as a feature filter or transformer only.

The \texttt{Pipeline} instance processes linearly the sequence of information
among its elements.  Its \texttt{fit!} implementation iteratively calls its
elements' \texttt{fit\_transform!} passing the output from one
\texttt{Computer} instance to the next \texttt{Computer} instance in the
sequence. Aside from sequential operations, two or more workflows can be
combined using the \texttt{ComboPipeline} instance.

The \texttt{fit!} and \texttt{transform!} functions for a \texttt{Learner} are
equivalent to training and prediction, respectively. A \texttt{Learner}
instance implements the algorithm to acquire the mapping between its input and
output during \texttt{fit!}  and applies the learned model to perform
prediction during \texttt{transform!}.

For a \texttt{Transformer}, \texttt{fit!} and \texttt{transform!} are
preprocessing operations to convert the training data into the same
mathematical or statistical transform.  Depending on the function used,
\texttt{fit!} can be a \texttt{noop} (no operation) like in \texttt{sqrt} or
\texttt{log} transform.  On the other hand, PCA or ICA uses \texttt{fit!} to
compute and store the coefficient matrix derived from its training input and
applies the same matrix in \texttt{transform!}.

\subsection{AMLP workflow}
\label{amlpworkflow}

The code in Listing \ref{lst:amlpcode} depicts the typical usage of the
AMLP toolkit. Lines 1--2 loads the AMLP package and the pro football
dataset~\cite{profb2014, openml}, respectively.  The aim is to predict whether
the game is held at \emph{home} or \emph{away} based on the following
(C)ategorical or (N)umerical features: FavoritePoints (N), UnderdogPoints
(N), Pointspread (N), FavoriteName (C), UnderdogName (C), Year (N), Week (N),
Weekday (C), and Overtime (C).

\begin{lstfloat}[htbp]
\begin{lstlisting}
using AMLP           # load package
(X,Y) = getprofb()   # load dataset

# Instantiate the pipeline elements
pca  = SKPreprocessor("PCA") 
mx   = SKPreprocessor("MinMaxScaler")  
ohe  = OneHotEncoder()
catf = CatFeatureSelector()  # categorical columns
numf = NumFeatureSelector()  # numerical columns
rf   = SKLearner("RandomForestClassifier")

# Setup the ML pipeline 
pipe = @pipeline ((catf |> ohe) + (numf |> mx |> pca)) |> rf

# train and predict
prediction = fit_transform!(pipe,X,Y) 
 
# compute avg accuracy by 10-fold cv
performance = crossvalidate(pipe,X,Y)  
\end{lstlisting}
  \caption{AMLP toolkit sample usage}
  \label{lst:amlpcode}
\end{lstfloat}

Lines 5--10 of Listing \ref{lst:amlpcode} create instances of the preprocessing
elements to be included in the pipeline.  The toolkit uses the
\emph{scikit-learn}~\cite{pedregosa2011} wrappers, \texttt{SKPreprocessor} and
\texttt{SKLearner}, to instantiate its transformers and learner: \texttt{PCA},
\texttt{MinMaxScaler}, and \texttt{RandomForestClassifier}. Other elements such
as: \texttt{OneHotEncoder}, \texttt{CatFeatureSelector}, and
\texttt{NumFeatureSelector} are implemented in pure \emph{Julia}.

In line 13 of Listing \ref{lst:amlpcode}, the expression:
\begin{lstlisting}[language = Julia,numbers=none]
   pipe = @pipeline ((catf |> ohe) + (numf |> mx |> pca)) |> rf
\end{lstlisting}
describes the preprocessing workflow of the input data with the
\texttt{RandomForest} classifier as the learner.  The expression,
\texttt{(x~|>~f)}, is equivalent to \texttt{f(x)} while the expression,
\texttt{(x~+~y)}, signifies feature union: \texttt{$\cup$(x,y)}.  The
expression, \texttt{(catf |> ohe)}, selects columns with categorical features
and transform them into one-hot representation. In \texttt{(numf |> mx |>
pca))}, numerical columns are selected then scaled by \texttt{MinMaxScaler} and
finally embedded in \texttt{PCA} subspace.  Both features are then concatenated
to become the input features of the \texttt{RandomForest} model (\texttt{rf}).

Line 13 of Listing \ref{lst:amlpcode} can also be written using
function calls as:
\begin{lstlisting}[language = Julia,numbers=none]
Pipeline(ComboPipeline(Pipeline(catf, ohe),
         Pipeline(Pipeline(numf, mx), pca)), rf)
\end{lstlisting}
This latter expression looks less understandable compared to the former. The
simplicity of AMLP pipeline expression becomes more significant in developing
AutoML algorithms or in data mining optimal pipelines among  different datasets
for surrogate modeling~\cite{avatar2020}.  More examples of AMLP usage
including its extensive documentation and source code can be found in its
open-source github resource [to be referenced if accepted].

\section{Pipeline search strategy benchmark}

The search for an optimal pipeline can be treated as a matching problem between a
group of learners and a group of preprocessing pipelines.  The objective is to
find the optimal pair of preprocessing pipeline and learner such that their
corresponding CV accuracy is the best among the rest of the
pairs.  We refer to any of these pairs as an ML pipeline (MLPL) in contrast to a
preprocessing pipeline (PRPL) where the last element is not a learner.  The
time complexity to search for all possible combinations of MLPL is dependent on
the number of learners as well as the size of elements in the PRPL:
\texttt{n(learners)~*~n(PRPL)}.

\subsection{Two-stage strategies}

To avoid the brute-force approach of exhaustive search, we attack the problem
by decomposing the search into two stages:

\begin{itemize}
  \item \emph{One-all} search strategy uses an arbitrarily
     chosen learner as the    engine of
    CV in searching for the best preprocessing pipeline performance
    in the first stage. The second stage proceeds by
    using the best pipeline found in
    the first stage to search for the best learner.
  \item \emph{All-one} search strategy uses an
    arbitrarily chosen pipeline as the base pipeline
    to search for the best
    learner during the first stage. The second stage
    uses the best learner found in the first
    stage to search for the best pipeline.
\end{itemize}

We call the first strategy \emph{one-all} to indicate the utilization of a
surrogate learner to evaluate all pipelines in the first stage.
Similarly, we call the second strategy \emph{all-one}
to indicate the evaluation of all learners under a
surrogate pipeline.  To aid the comparison, we also implemented the
\emph{all-all} strategy which is an exhaustive search of the performance of all
combinations of learners and PRPL to get the best MLPL.

In our implementation, we call the pipeline consisting of scaler (sc) and
feature extractor (fx) a one-block PRPL, expressed as: \texttt{(sc |> fx)}.
In the experiments, we use the maximum of two-block PRPL,
\texttt{((sc1~|>~fx1)~+~(sc2~|>~fx2))}, where \texttt{sc1~!=~sc2} or
\texttt{fx1~!=~ fx2}.  The pipeline can easily be extended by adding more PRPL
blocks depending on the complexity of the dataset at the expense of longer
computation time.

\subsection{Experimental setup}

\begin{table*}[htbp]
  \caption{OpenML Datasets}
  \label{tab:openmldata}
  \centering
  \tiny
  \begin{tabular}{{@{} rccccccccc @{}}}
    \toprule
	 Dataset      & NAs  & Size & Cat & Num & Rows  & Cols & Class & Best Result       & AutoML \\
    \midrule
	 analcatdata  & 0    & 22K  & 5   & 0   & 9873  & 5    & 6     & 75.10 $\pm$ 02.59 & ATM \\
    breast-w     & 16   & 18K  & 1   & 9   & 699   & 10   & 2     & 01.44 $\pm$ 02.22 & ATM \\
    cmc          & 0    & 30K  & 8   & 2   & 1473  & 10   & 3     & 43.33 $\pm$ 01.52 & TPOT \\
    credit-app   & 67   & 39K  & 10  & 6   & 690   & 16   & 2     & 11.11 $\pm$ 03.52 & ATM \\
    eucalyptus   & 448  & 74K  & 6   & 14  & 736   & 20   & 5     & 35.41 $\pm$ 05.02 & AutoSk \\
	 first-order  & 0    & 2.6M & 1   & 51  & 6118  & 52   & 6     & 38.48 $\pm$ 01.95 & TPOT \\
    GestureSeg   & 0    & 3.3M & 1   & 32  & 9873  & 33   & 5     & 32.78 $\pm$ 03.59 & TPOT \\
    jm1          & 25   & 826K & 1   & 21  & 10885 & 22   & 2     & 18.13 $\pm$ 01.55 & TPOT \\
    profb        & 1200 & 25K  & 5   & 5   & 672   & 10   & 2     & 32.43 $\pm$ 05.84 & TPOT \\
	 plants-shape & 0    & 892K & 1   & 64  & 1600  & 65   & 100   & 36.54 $\pm$ 03.53 & AutoSk \\
	 sick         & 6064 & 300K & 23  & 7   & 3772  & 30   & 2     & 01.30 $\pm$ 00.87 & TPOT \\
	 soybean      & 2337 & 171K & 36  & 0   & 683   & 36   & 19    & 06.59 $\pm$ 02.79 & ATM \\
    \bottomrule
  \end{tabular}
\end{table*}

Table \ref{tab:openmldata} summarizes the statistical features of the 12 OpenML
datasets~\cite{openml} used in the experiment.  Also included are the best
results in a 4-hour time budget  discussed in the review paper
of \cite{zoller2019}.  The comparison involves 5 AutoML approaches, namely:
TPOT~\cite{tpot2016}, Auto-Sklearn~\cite{Feurer:AML2019Chp6,Feurer:NIPS2015},
ATM~\cite{atm2017}, Hyperopt-Sklearn~\cite{bergstra2011nips,bergstra2012}, and
Random Search~\cite{anderson1953}.  Among these approaches, TPOT and ATM have
dominated the rankings in terms of best performance based on the
CV classification errors on the 12 datasets as shown in the last two columns.

The AMLP experiments use the following transformers and learners including
\texttt{Noop} (no operation):
\begin{itemize}
  \item 6 scalers: Standard, MinMax, Robust, Normalizer, PowerTransfomer, Noop
  \item 4 feature extractors: PCA, ICA, FactorAnalysis, Noop
  \item 6 learners: RandomForest, AdaBoost, DecisionTree, GradientBoosting, LinearSvm, RbfSvm
\end{itemize}

For the one-block PRPL experiment, the size of all MLPL  for the exhaustive
search will be $6\times4\times6 = 144$.  A relatively small size that can
easily be evaluated to search for the optimal solution which is ideal for
experimental comparison with the two-stage strategies. However, the size of all
combinations of MLPL suddenly explodes to more than 3,000 pipelines by just
adding another PRPL block. With 10-fold CV per MLPL,
the exhaustive search requires 1440 CV operations
for a one-block MLPL and more than 30,000 CV operations for a two-block MLPL.

The \emph{one-all} search strategy requires $6\times4 = 24$ one-block PRPL
matched to one learner for the first stage. The second stage requires matching
the best pipeline against 6 learners.  In this strategy, a total of 30 MLPL
pipelines are needed to get the optimal or suboptimal solution.  With 10-fold
CV per MLPL, \emph{one-all} requires 300 CV operations which is 4.8x smaller
compared to the 1440 CV operations needed for the exhaustive search strategy.

For the \emph{all-one} strategy, 6 learners are matched to a single pipeline in the
first stage. The best learner found is then matched to $6\times4 = 24$ PRPL in
the second stage.  In total, the \emph{all-one} uses 30 MLPL which translates
to 300 CV operations similar to the \emph{one-all}. The actual
runtime performance for \emph{one-all} and \emph{all-one} in the same
dataset depends mostly on the best learner performance in
\emph{all-one} and the surrogate learner performance in \emph{one-all}.

All datasets undergo common cleaning workflow in all experiments.
The entire process is summarized in the following:
\begin{lstlisting}[language = Julia,numbers=none]
basepipeline = @pipeline colnarm |> rownarm |> ((catf  |> ohe) + numf)
\end{lstlisting}

For each dataset, column-wise removal (\texttt{colnarm})
of missing values (\emph{NA}) is carried out on
those columns with \emph{NA} count
greater than 10\% of the row size followed by
row-wise removal (\texttt{rownarm})
of any remaining \emph{NA}s.
After \emph{NA} removal, categorical features (\texttt{catf}) of the
dataset are transformed to one-hot (\texttt{ohe})
representation and concatenated with its numerical
features (\texttt{numf}).

\subsection{Results and Discussion}

\begin{table*}[htbp]
  \caption{\emph{All-all} (one-block) search strategy}
  \label{tab:singleblock}
  \centering
  \tiny
  \begin{tabular}{{@{} rcrrrlll @{}}}
    \toprule
	 & & & & & \multicolumn{2}{c}{Block} \\
    \cmidrule(r){6-7}
	 Dataset          & Rank & AvgErr & Std  & Time & Sc         & Fx   & Learner \\
    \midrule
    analcatdata-dmf  & 2    & 76.52  & 4.92 & 0.13 & normalizer & noop & rbfsvc\\
    breast-w         & 2    & 02.34  & 1.71 & 0.08 & minmax     & noop & rbfsvc\\
    cmc              & 1    & 42.57  & 3.37 & 0.18 & noop       & noop & gb\\
    credit-approval  & 2    & 11.42  & 2.60 & 0.09 & normalizer & noop & rf\\
    eucalyptus       & 3    & 36.02  & 6.61 & 0.17 & noop       & noop & gb\\
	 first-order-theo & 1    & 36.61  & 1.44 & 1.27 & noop       & noop & rf\\
    GestureSeg       & 1    & 31.96  & 1.28 & 1.87 & minmax     & noop & rf\\
    jm1              & 1    & 17.97  & 0.61 & 1.84 & stdsc      & noop & rf\\
    profb            & 1    & 25.59  & 5.15 & 1.84 & stdsc      & noop & lsvc\\
    plants-shape     & 1    & 30.25  & 3.03 & 4.29 & powertf    & pca  & rf\\
    sick             & 4    & 06.40  & 0.70 & 0.41 & noop       & noop & gb\\
    soybean          & 5    & 14.62  & 2.48 & 0.27 & robustsc   & noop & lsvc\\
	 \cmidrule{2-5}
	 Median           & 1.5  & 27.92  & 2.54 & 0.23 &            &      & \\
    \bottomrule
  \end{tabular}
\end{table*}

Table \ref{tab:singleblock} summarizes the performance of \emph{all-all}
strategy. The metric of performance is based on the average classification
error (\emph{AvgErr}) using 10-fold CV. The ranking is based on its
\emph{AvgErr} performance in comparison to the 5 AutoML approaches~\cite{zoller2019} summarized in Table~\ref{tab:openmldata}.
The \emph{all-all} median overall rank is 1.5 using one-block pipeline.
Its median runtime to
perform one-block exhaustive search is 0.23 hour or 13.8 minutes per dataset.
This result is encouraging because the median runtime is significantly less
compared to the 5 AutoML algorithms with a 4-hour budget.  Closely examining 
the runtime of each dataset, \emph{plants-shape} dataset
requires 4.29 hours to finish while
the rest require at most 1.87 hours to run. Take note, however, that the
comparison for runtime is not a fair comparison, as we did not have the time to repeat the
experiments done in ~\cite{zoller2019}
using the same machine with our two-stage strategies. Our experiments were conducted using 2017 Model of MacBook Pro with 2.8 GHz Quad-Core Intel Core i7 and 16 GB of RAM.

The solution can be improved further by searching all two-block pipelines,
but the size of the search space becomes a limiting factor. Ideally, if either
\emph{one-all} or \emph{all-one} has similar optimal results with that of
\emph{all-all}, the runtime required will be significantly less to attack
the two-block MLPL.  The succeeding discussions of the results of
experiments aim to find out if there is a significant runtime
saving in either one or both of the two-stage strategies based
on their performance relative to the exhaustive \emph{all-all}
strategy.

\begin{table*}[htbp]
  \centering
  \caption{\emph{One-all} (one-block)}
  \label{tab:oneblock-skrf}
  \tiny
  \begin{tabular}{{@{} rcrrrlll @{}}}
	 \toprule
	 & & & & & \multicolumn{2}{c}{Block} \\
	 \cmidrule(r){6-7}
	 Dataset          & Rank & AvgErr & Std  & Time & Sc       & Fx    & Lr \\
	 \midrule
	 analcatdata-dmf  & 3    & 78.42  & 5.77 & 0.35 & minmax   & factA & ada\\
	 breast-w         & 3    & 2.78   & 1.88 & 0.02 & stdsc    & pca   & rbfsvc\\
	 cmc              & 4    & 45.76  & 2.94 & 0.02 & norm     & noop  & gb\\
	 credit-approval  & 2    & 11.34  & 4.98 & 0.09 & norm     & noop  & rf\\
	 eucalyptus       & 3    & 36.79  & 7.61 & 0.09 & noop     & noop  & rf\\
	 first-order-theo & 1    & 36.89  & 1.68 & 0.32 & powertf  & noop  & rf\\
	 GestureSeg       & 1    & 32.04  & 1.16 & 0.15 & noop     & noop  & rf\\
	 jm1              & 1    & 18.01  & 0.51 & 0.34 & powertf  & noop  & rf\\
	 profb            & 1    & 28.87  & 4.86 & 0.09 & noop     & ica   & rf\\
	 plants-shape     & 1    & 30.80  & 4.44 & 0.13 & powertf  & pca   & rf\\
	 sick             & 4    & 6.60   & 0.66 & 0.06 & robustsc & pca   & rf\\
	 soybean          & 5    & 15.29  & 4.78 & 0.34 & robustsc & noop  & lsvc\\
	 \cmidrule{2-5}
	 Median           & 2.50 & 29.84  & 3.69 & 0.11 &          &       & \\
	 \bottomrule
  \end{tabular}
\end{table*}

\begin{table*}[htbp]
  \centering
  \caption{\emph{All-one} (one-block)}
  \label{tab:all-one-single}
  \tiny
  \begin{tabular}{{@{} rcrrrlll @{}}}
	 \toprule
	 & & & & & \multicolumn{2}{c}{Block} \\
	 \cmidrule(r){6-7}
	 Dataset          & Rank & AvgErr & Std  & Time & Sc       & Fx   & Lr \\
	 \midrule
	 analcatdata-dmf  & 3    & 77.92  & 3.42 & 0.01 & noop     & noop & rbfsvc\\
	 breast-w         & 2    & 2.34   & 2.08 & 0.01 & robustsc & noop & rbfsvc\\
	 cmc              & 1    & 42.36  & 4.14 & 0.08 & stdsc    & noop & gb\\
	 credit-approval  & 2    & 11.26  & 3.72 & 0.03 & norm     & noop & rf\\
	 eucalyptus       & 3    & 36.44  & 4.60 & 0.04 & robustsc & noop & rf\\
	 first-order-theo & 1    & 36.48  & 2.08 & 0.17 & powertf  & noop & rf\\
	 GestureSeg       & 1    & 32.39  & 1.09 & 0.30 & robustsc & noop & rf\\
	 jm1              & 1    & 18.08  & 0.96 & 0.45 & stdsc    & noop & rf\\
	 profb            & 1    & 25.14  & 6.42 & 0.02 & robustsc & noop & lsvc\\
	 plants-shape     & 1    & 30.35  & 3.86 & 0.12 & stdsc    & pca  & rf\\
	 sick             & 4    & 6.54   & 0.49 & 0.07 & robustsc & pca  & gb\\
	 soybean          & 5    & 14.66  & 3.96 & 0.01 & stdsc    & noop & rbfsvc\\
	 \cmidrule{2-5}
	 Median           & 1.5  & 27.75  & 3.57 & 0.05 &          &      & \\
	 \bottomrule
  \end{tabular}
\end{table*}

\begin{table*}[htbp]
  \caption{\emph{All-one} (two-block)}
  \label{tab:twoblocks-all-one}
  \centering
  \tiny
  \begin{tabular}{{@{} rcrrr|ll|ll|l @{}}}
	 \toprule
	 & & & & & \multicolumn{2}{|c|}{Block 1} & \multicolumn{2}{c|}{Block 2} \\
	 \cmidrule(r){6-9}
	 Dataset      & Rank & AvgErr & Std  & Time  & Sc1      & Fx1   & Sc2      & Fx2   & Lr\\
	 \midrule
	 analcatdata  & 2    & 75.54  & 4.91 & 0.56  & minmax   & noop  & powertf  & factA & ada\\
	 breast-w     & 2    & 2.11   & 1.65 & 0.41  & norm     & noop  & minmax   & noop  & rbfsvc\\
	 cmc          & 1    & 42.57  & 3.45 & 2.28  & stdsc    & noop  & robustsc & noop  & gb\\
	 credit-app   & 1    & 11.03  & 4.59 & 0.50  & stdsc    & noop  & powertf  & factA & lsvc\\
	 eucalyptus   & 1    & 33.40  & 6.13 & 2.78  & stdsc    & noop  & noop     & factA & gb\\
	 first-order  & 1    & 36.37  & 2.49 & 5.51  & noop     & ica   & noop     & noop  & rf\\
	 GestureSeg   & 1    & 30.46  & 1.03 & 9.69  & norm     & ica   & stdsc    & noop  & rf\\
	 jm1          & 1    & 17.67  & 0.79 & 16.62 & robustsc & noop  & norm     & pca   & rf\\
	 profb        & 1    & 24.69  & 5.29 & 0.61  & stdsc    & ica   & stdsc    & noop  & lsvc\\
	 plants-shape & 1    & 26.05  & 2.77 & 3.97  & stdsc    & factA & norm     & pca   & rf\\
	 sick         & 4    & 6.26   & 1.01 & 2.65  & robustsc & pca   & noop     & noop  & gb\\
	 soybean      & 5    & 13.24  & 2.62 & 0.57  & robustsc & noop  & noop     & noop  & lsvc\\
	 \midrule
	 Median/Mode  & 1.0  & 25.37  & 2.70 & 2.62  & stdsc    & noop  & noop     & noop  & rf\\
	 \bottomrule
  \end{tabular}
\end{table*}

\begin{table*}[htbp]
  \caption{\emph{All-one with Surrogates}}
  \label{tab:surrogate}
  \centering
  \tiny
  \begin{tabular}{{@{} r|crrr|crrr|crrr| @{}}}
	 \toprule
	              & \multicolumn{4}{c|}{Baseline: All-one} & \multicolumn{4}{c|}{Surrogate: Baseline+PRP} & \multicolumn{4}{c|}{Surrogates: Baseline+PRP+LR} \\
	 \cmidrule(r){2-13}
	 Dataset      & Rank & AvgErr & Std  & Time   & Rank & AvgErr & Std  & Time   & Rank & AvgErr & Std  & Time \\
	 \midrule                                                                                                   
	 analcatdata  & 2    & 75.54  & 4.91 & 0.56   &  2   & 76.67  & 4.24 & 0.07   &  2   & 76.17  & 0.92 & 0.00 \\
	 breast-w     & 2    & 02.11   & 1.65 & 0.41  &  2   & 02.49  & 1.70 & 0.02   &  2   & 02.33  & 2.49 & 0.00 \\
	 cmc          & 1    & 42.57  & 3.45 & 2.28   &  1   & 42.84  & 4.64 & 0.07   &  2   & 44.13  & 3.68 & 0.01 \\
	 credit-app   & 1    & 11.03  & 4.59 & 0.50   &  2   & 11.60  & 4.83 & 0.03   &  2   & 12.08  & 3.80 & 0.04 \\
	 eucalyptus   & 1    & 33.40  & 6.13 & 2.78   &  3   & 36.37  & 7.03 & 0.16   &  2   & 35.63  & 6.27 & 0.22 \\
	 first-order  & 1    & 36.37  & 2.49 & 5.51   &  1   & 36.73  & 2.18 & 0.13   &  1   & 36.63  & 2.07 & 0.28 \\
	 GestureSeg   & 1    & 30.46  & 1.03 & 9.69   &  1   & 31.58  & 1.45 & 0.44   &  5   & 45.19  & 0.96 & 0.76 \\
	 jm1          & 1    & 17.67  & 0.79 & 16.62  &  1   & 18.01  & 0.86 & 0.23   &  1   & 17.98  & 1.47 & 0.40 \\
	 profb        & 1    & 24.69  & 5.29 & 0.61   &  1   & 24.86  & 6.02 & 0.02   &  1   & 24.41  & 4.42 & 0.04 \\
	 plants-shape & 1    & 26.05  & 2.77 & 3.97   &  1   & 32.06  & 2.93 & 0.17   &  3   & 44.31  & 5.53 & 0.61 \\
	 soybean      & 5    & 13.24  & 2.62 & 0.57   &  5   & 14.29  & 3.55 & 0.02   &  5   & 14.29  & 4.83 & 0.03 \\
	 \midrule                                                                                                   
	 Median       & 1.0  & 26.05  & 2.77 & 2.28   & 1.0  & 31.58  & 3.55 & 0.07   &  2.0 & 35.63  & 3.68 & 0.04 \\
	 \bottomrule
  \end{tabular}
\end{table*}

Tables \ref{tab:oneblock-skrf} and \ref{tab:all-one-single} show the
performance of \emph{one-all} and \emph{all-one} strategies, respectively, for one-block MLPL.
Their ranking is based on their performances against the 5 AutoML algorithms.
Table \ref{tab:oneblock-skrf} summarizes the performance of \emph{one-all}
using the \texttt{RandomForest} surrogate.
All other learners tested as surrogate have similar 2.0 to
2.5 median rank. The \emph{all-one} strategy in Table \ref{tab:all-one-single}
uses the pipeline expression,\texttt{(((catf~|>~ohe)~+~numf))~|>~robustsc},
as the surrogate pipeline to
search for the best learner in the first stage.  The expression indicates
one-hot encoding~(ohe) of the categorical features~(\texttt{catf}) combined
with the numerical features~(\texttt{numf}) and transformed by robust scaling~(\texttt{robustsc}).

Comparing Tables \ref{tab:singleblock}, \ref{tab:oneblock-skrf} and \ref{tab:all-one-single}, 
the \emph{all-one} strategy median rank of 1.5 is
similar to \emph{all-all} and significantly better than
the 2.5 median of \emph{one-all}. The \emph{all-one}
median validation error of 27.75
$\pm$ 3.69 is also slightly better than \emph{all-all} (27.92$\pm$ 2.54).
While both have similar validation errors, the biggest advantage of
\emph{all-one} is its speed which is significantly faster than \emph{all-all}.
Its median time duration for all 12 datasets is just 0.05
hour (3 minutes) compared to the runtime median of 0.23 hour (13.8 minutes)
for \emph{all-all}. Furthermore, the dataset with worst runtime is the
\emph{analcadata-dmf} which requires 0.35 hour or 21 minutes.  This worst
runtime is still significantly less than the 4-hour budget allotted to the 5
AutoML algorithms. It is interesting to note that most solutions in
\emph{all-one} do not employ any feature extraction but only scaling. We
can consider the solutions in \emph{all-one} to be sparsed relative to
the dominant presence of \texttt{noop} in the feature extraction part
of the one-block MLPL.

Inspired by these promising results, the next experiment applies the
\emph{all-one} strategy to two-block MLPL with the results summarized in Table
\ref{tab:twoblocks-all-one}. The \emph{all-one} strategy achieves a 1.0 median
rank which suggests that its performance is at least as good or better than the
best results of the 5 AutoML algorithms with respect to the 12 datasets.
While the runtime median of \emph{all-one} is
2.62 hours, 3 datasets require more than 4 hours to finish.  
In the future, it will be interesting to incorporate the time budget into the 
\emph{all-one} strategy and run the other AutoMLs in the same machine to have fair comparison.

Using the \emph{all-one} strategy, there is a trade-off in its applications
to one-block or two-block pipelines. The former has significantly quick runtime
but may not be optimal, while the latter may require relatively longer runtime
than the former for some datasets but with a higher likelihood of being optimal.

Another interesting observation in Tables \ref{tab:singleblock},
\ref{tab:all-one-single}, and \ref{tab:twoblocks-all-one}
regarding the composition of their best solutions
is the high occurrence of \texttt{noop}.
We can consider their solutions to be sparsed because in many cases
they do not fully utilize
the 2 out 5 non-noop scalers and 2 out of 3 non-noop feature extractors in their optimal
pipelines.  This insight can be valuable by using the
datamined signatures of the optimal
solutions to predict structure complexity of the pipeline for an efficient search.
By mapping data metafeatures and the corresponding optimal pipeline signatures,
we can train a metalearner to guide which subset of pipelines or elements
of the pipelines can be used as a starting point in search.

Table~\ref{tab:surrogate} summarizes the results of implementing these
insights using PRP and LR surrogate models.
The PRP-surrogate is trained to learn the mapping between 
dataset metafeatures and its corresponding optimal pipeline 
signature complexity. Pipeline complexity has 4 categories based
on the presence of \emph{noop}. Category 1 has zero or one \emph{noop}, 
category 2 has two \emph{noops}, etc. More \emph{noops} imply 
less pipeline complexity. On the other hand, the LR-surrogate
is used to learn the mapping between the dataset metafeatures 
with its corresponding optimal learner type: Ensemble vs SVM. Both
surrogate models are trained using the OpenML-CC18~\cite{openmlcc18} 
datasets by extracting their metafeatures using \emph{pymfe}~\cite{pymfe2020}. The extracted datasets, results, and julia-based pseudocode for surrogate modeling can be found in the supplementary material submission of the paper.

We use the \emph{all-one} strategy as the baseline and extended it with 
PRP and LR surrogates to find out their effect in prediction error
and computation time. While there is a 5\% increase in error by
using PRP-surrogate, the median computation time went down from 2.28
hours to just 4.2 minutes. In spite of the 5\% drop in accuracy,
PRP-surrogate median rank of 1.0 still indicates superior performance
relative to other AutoML approaches.

The exponential reduction in computation time of PRP-surrogate 
is due to the use of smaller search space due to the removal of unnecessary 
preprocessing elements. Incorporating further the LR-surrogate
increases the median prediction error by another 4\% 
but reduces the median computation time to just 2.4 minutes. 
The main trade-off in relying on more surrogates is the reduction of 
computation time at the expense of less accurate prediction. 
Depending on the requirements, using one or more surrogates 
can be necessary to speed-up computation if the corresponding drop in 
prediction accuracy is acceptable which can be true in some 
application domains.

Among the datasets, the two-stage strategies have consistent low performance in
both \emph{sick} and \emph{soybean} problems.  Both datasets are characterized
by large number of missing values compared to the rest of the datasets.
There are 6064 \emph{NA}s in \emph{sick} and 2337 \emph{NA}s in
\emph{soybean}. The poor performance of the two-stage strategies
can be attributed to the non-implementation of imputation or
interpolation filter in the \texttt{basepipeline} data
cleaning routine. Future experiments
will examine which imputation or interpolation routines
can be used to improve AMLP solutions.

\section{Conclusion}

Based on the set of experiments we conducted, 
the \emph{all-one} strategy provides the optimal solution in 
a significantly shorter duration relative to other
AutoML algorithms. The results indicate that using a simple pipeline
in the first stage consisting of one-hot encoded categorical 
features together with their numerical features
under robust scaling provides a good representation of the dataset difficulty
for evaluating the best matched optimal learner from the group of learners.
The winning learner picked by the first stage can then be used to improve the
solution further by looking for a more optimal match with the rest of the
pipelines. The \emph{all-one} strategy computation speed can be exponentially 
reduced by utilizing PRP and LR surrogates in exchange of lower accuracy
which remains competitive relative to other AutoML approaches.

The easy and straightforward experimental setup in the implementation of these
series of experiments can be attributed to the 
usability of \emph{Julia} and AMLP toolbox.
AMLP's support for a high-level description of the pipelines
makes it easy to track which of the
pipeline elements are dominant among the optimal solutions. These insights can
be used in the development of the runtime search strategy in future algorithms.
Data mining optimal pipelines become much easier because one can
directly use these high-level expressions to perform text-mining,
frequency pattern mining, text associations,
and NLP on the elements and structure of the optimal pipeline solutions.

The high-level and easy composability in AMLP can result in the creation of
complex pipelines that can barely be understood. For similar reasons that
regularization and information criterion are employed in mathematical and
statistical modeling, any AutoML implementation relying on AMLP and similar
toolkits must incorporate strategies that take into account a good balance
between pipeline complexity, accuracy, and explainability for easy scrutiny,
comprehensibility, and verification bias of its solutions.

\bibliography{amlp}
\bibliographystyle{unsrt}

\end{document}